\documentclass[letterpaper, 10 pt, conference]{ieeeconf}

\IEEEoverridecommandlockouts
\usepackage[utf8]{inputenc}
\usepackage{amsmath,amssymb,amsfonts}
\usepackage{graphicx}
\usepackage{cite}
\usepackage{booktabs}
\usepackage[hidelinks,hypertexnames=false]{hyperref}

\usepackage{textcomp}
\usepackage{xcolor}
\usepackage{threeparttable}
\usepackage{tabularx}
\usepackage{array}

\newcolumntype{Y}{>{\raggedright\arraybackslash}X}

\usepackage{float}
\floatstyle{ruled}
\newfloat{algorithm}{tbp}{loa}
\floatname{algorithm}{Algorithm}
\newcounter{algln}
\newenvironment{pseudo}{%
  \begin{list}{\scriptsize\ttfamily\arabic{algln}:}{%
    \usecounter{algln}%
    \setlength{\itemsep}{0pt}\setlength{\parsep}{0pt}\setlength{\topsep}{2pt}%
    \setlength{\leftmargin}{1.7em}\setlength{\labelwidth}{1.2em}\setlength{\labelsep}{0.4em}}%
  \footnotesize}{\end{list}}
\newcommand{\kw}[1]{\textbf{#1}}
\newcommand{\IND}{\hspace*{1.1em}}

\def\IEEEkeywordsname{Keywords}
\newenvironment{IEEEkeywords}{\par\noindent\textbf{\IEEEkeywordsname}\ \ignorespaces}{\par}

\begin{document}
\bstctlcite{RoboSegBSTcontrol}

\title{RoboSeg: Online Part-Level Semantic Reconstruction for Robotic Manipulation via a Single Eye-in-Hand Camera}

\author{Zhaochen Lan$^{1}$ and Mengxiang Lin$^{1,\dagger}$% <-this % stops a space
\thanks{$^{1}$Zhaochen Lan and Mengxiang Lin are with the School of Mechanical Engineering and Automation, Beihang University, Beijing 100191, China
        {\tt\small \{lanzhaochen, linmx\}@buaa.edu.cn}}%
\thanks{$^{\dagger}$Corresponding author: Mengxiang Lin {\tt\small (linmx@buaa.edu.cn)}.}%
}

\maketitle

\begin{abstract}
Robotic manipulation requires perception systems that identify actionable parts such as handles, rims, triggers, and tool tips, not merely object categories or point clouds.
This paper presents \textbf{RoboSeg}, a part-level semantic reconstruction system that links vision-language model (VLM) functional-part discovery, asynchronous online RGB-D semantic reconstruction, and task-oriented grasp generation without requiring CAD models or pre-scanned meshes.
RoboSeg queries a VLM on the initial RGB observation to obtain compact functional part prompts, then scans with two asynchronous streams: a high-frequency geometry thread for RGB-D odometry and truncated signed distance function (TSDF) fusion, and a keyframe-triggered semantic thread for SAM3~\cite{carionSAM3Segment2025} part masks.
Projected masks are fused by voxel-level temporal voting into a persistent part-labeled point cloud; RoboSeg uses this map to assign AnyGrasp~\cite{fangAnyGraspRobustEfficient2023} 6-DoF candidates to semantic parts and select grasps consistent with the task-relevant part label.
RoboSeg reaches 83.4\% mean part intersection-over-union (mIoU) over manually labeled objects; in a 24-trial physical pilot across four objects and eight tasks, the selected grasp contacts the requested part in all trials and achieves 21/24 combined task successes.
These results characterize RoboSeg as a semantic indexing layer for task-conditioned manipulation, with AnyGrasp retained as the proposal generator.
\end{abstract}

\begin{IEEEkeywords}
Robotic Manipulation, Semantic Mapping, Vision-Language Models, Part Segmentation, Grasping
\end{IEEEkeywords}

\section{Introduction}

Effective robotic manipulation of everyday objects requires reasoning about two coupled questions: (i)~which regions of an object afford stable contact, and (ii)~how the selected contact region supports the intended task.
For many objects, the operationally meaningful unit is not the object category as a whole, but a functional part: a mug handle affords lifting, a watering-can spout enables controlled pouring, and a screwdriver handle plays a different mechanical role from its shaft.
This requirement leads to a problem setting distinct from conventional open-vocabulary 3D part segmentation.

Most 3D part-level segmentation methods, including PartSLIP~\cite{liuPartSLIPLowShotPart2023,zhouPartSLIPEnhancingLowShot2023}, PartSTAD~\cite{kimPartSTAD2Dto3DPart2024}, and ZeroPS~\cite{xueZeroPSHighqualityCrossmodal2025}, assume known or precomputed complete object geometry, such as a CAD model, mesh, or point cloud, and optimize part labels without closing the loop to manipulation.
A robot must instead build and use the representation while observing an unknown workspace object.

\begin{figure}[t]
    \centering
    \includegraphics[width=\linewidth]{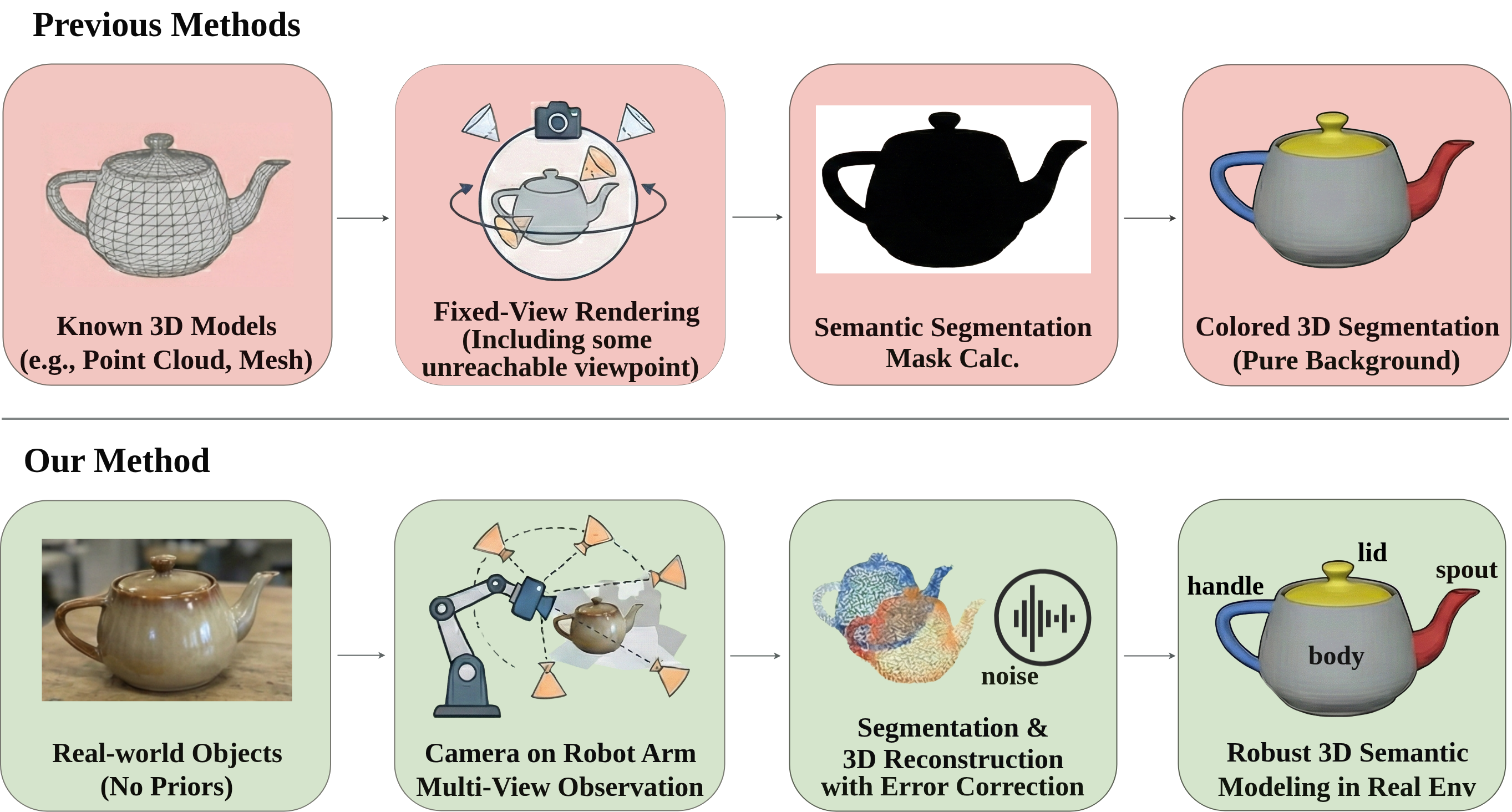}
    \caption{\textbf{Problem setting contrast.} Offline 3D part segmentation typically starts from known or precomputed complete object geometry, such as a mesh or complete point cloud, and then outputs part labels. RoboSeg targets an online manipulation setting: the robot observes an unknown object with an eye-in-hand RGB-D camera, discovers functional parts from the initial view, reconstructs geometry during scanning, and maintains a part-level map for grasp-candidate selection.}
    \label{fig:comparison}
\end{figure}

RoboSeg starts from raw RGB-D observations captured by an eye-in-hand camera and discovers manipulation-relevant parts while building the 3D model used by the robot.
Modern 2D vision foundation models, including Segment Anything (SAM)~\cite{kirillovSegmentAnything2023} and Grounding DINO~\cite{liuGroundingDINOMarrying2024}, can segment arbitrary visual regions, but their unconstrained use in this online setting exposes two practical gaps.
First, segmentation is often driven by texture, color, or specular boundaries rather than manipulation-relevant structure.
Second, 2D masks must be lifted into a stable 3D representation despite camera motion, pose drift, and partial observations.

Open-vocabulary semantic mapping systems~\cite{jatavallabhulaConceptFusionOpensetMultimodal2023,guConceptGraphsOpenVocabulary3D2024,keethaSplaTAMSplatTrack2024} lift language-aligned visual features into 3D maps, but usually target room-scale or object-level scene understanding.
For manipulation, the decisive question is whether the map supports task-relevant grasp selection.

RoboSeg addresses this gap by integrating open-vocabulary part reasoning, asynchronous online 3D semantic mapping, and task-conditioned grasp-candidate selection in a single pipeline.
The system uses a UR5 manipulator with an eye-in-hand Intel RealSense D435i camera.
At initialization, Qwen3-VL~\cite{yangQwen3TechnicalReport2025} analyzes the first RGB frame and returns a compact list of self-contained part names, such as \textit{watering can handle} or \textit{screwdriver shaft}.
During scanning, an asynchronous architecture decouples high-frequency RGB-D odometry and TSDF fusion from lower-frequency, keyframe-triggered SAM3~\cite{carionSAM3Segment2025} semantic inference.
The semantic masks are projected into the TSDF volume and fused by a voxel voting tensor that suppresses frame-to-frame label flicker.
After scanning, the system exports a part-labeled point cloud and instance table, runs AnyGrasp~\cite{fangAnyGraspRobustEfficient2023} on the reconstructed geometry, and indexes each grasp candidate by semantic part so that a task can select among part-specific candidates rather than relying only on an object-level grasp list.

This paper makes two contributions.
First, it presents an online perception-to-grasp-candidate interface that connects VLM-guided functional part discovery, asynchronous 3D semantic reconstruction, and AnyGrasp-based part-indexed candidate selection without CAD models, object templates, or pre-scanned meshes.
Second, it evaluates this interface through three questions: whether the discovered parts match manipulation-relevant annotations and remain comparable to offline methods under aligned labels, whether the geometric backbone is accurate enough for candidate indexing, and whether the resulting map can guide real task-part grasp attempts.

The evaluation is organized around this interface rather than a single benchmark score: each experiment isolates one link in the perception-to-grasp pipeline and states the evidence it provides.

\section{Related Work}

\subsection{Part Segmentation and Vision Foundation Models}
Part-level reasoning has a long history that predates vision-language models.
Deformable part-based detectors~\cite{felzenszwalbObjectDetectionDiscriminatively2010}, part-based networks for fine-grained recognition~\cite{zhangPartbasedRCNNsFinegrained2014}, densely part-annotated benchmarks~\cite{chenDetectWhatYou2014}, and early deep networks for semantic part segmentation~\cite{tsogkasDeepLearningSemantic2015} established object parts as a recognition primitive well before large-scale vision-language pretraining, and open-vocabulary part segmentation such as VLPart~\cite{sunGoingDenserOpenVocabulary2023} later extended part prediction to arbitrary part vocabularies without per-part supervision.
RoboSeg builds on this line of work rather than treating part-level perception as a capability introduced by VLMs.
More recently, SAM~\cite{kirillovSegmentAnything2023}, SAM3~\cite{carionSAM3Segment2025}, Grounding DINO~\cite{liuGroundingDINOMarrying2024}, and Grounded-SAM~\cite{renGroundedSAMAssembling2024} enable open-vocabulary 2D segmentation from point, box, concept, or text prompts.
These models are attractive for robotics because they reduce the need for task-specific mask annotation.
Their raw outputs, however, are not automatically suitable for manipulation.
In ``segment everything'' mode, they often partition an object according to local appearance, while a robot requires a smaller set of functional components.
Recent VLMs such as Qwen3-VL~\cite{yangQwen3TechnicalReport2025} and GPT-4V-style robotic reasoning systems~\cite{wakeGPT4VisionRoboticsMultimodal2024a} can infer object affordances from images.
RoboSeg uses this capability as a front-end constraint: VLM reasoning selects the part vocabulary before segmentation, rather than asking the segmentation model to discover all visible fragments.
Offline 3D part methods such as PartSLIP, PartSTAD, SAMPart3D, and FindAnyPart~\cite{liuPartSLIPLowShotPart2023,kimPartSTAD2Dto3DPart2024,yangSAMPart3DSegmentAny2024,maFindAnyPart2025} lift part segmentation to three dimensions and serve as useful baselines, but they assume complete object geometry, typically a mesh or point cloud prepared before manipulation, and do not maintain an online perception-action loop.

\subsection{Online 3D Reconstruction and Semantic Mapping}
Volumetric fusion with truncated signed distance functions (TSDFs) remains a practical representation for dense geometry~\cite{curlessVolumetricMethodBuilding1996}.
Recent SLAM and mapping systems, including Vox-Fusion~\cite{yangVoxFusionDenseTracking2022}, Co-SLAM~\cite{wangCoSLAMJointCoordinate2023}, ConceptFusion~\cite{jatavallabhulaConceptFusionOpensetMultimodal2023}, ConceptGraphs~\cite{guConceptGraphsOpenVocabulary3D2024}, and SplaTAM~\cite{keethaSplaTAMSplatTrack2024}, show that geometry and open-vocabulary semantics can be integrated at scale.
Tabletop manipulation imposes a different operating regime.
The camera remains close to the object, the viewpoint changes continuously with wrist motion, the target geometry is only partially observed, and the semantic granularity must be part-level.
These constraints motivate the online, eye-in-hand part-level mapping that RoboSeg develops, in contrast to the offline 3D part pipelines discussed above.

\subsection{Task-Oriented Grasping}
Learning-based grasp detectors such as AnyGrasp~\cite{fangAnyGraspRobustEfficient2023} and benchmarks such as GraspNet-1Billion~\cite{fangGraspNet1BillionLargeScaleBenchmark2020} provide dense 6-DoF grasp candidates from point clouds.
Language and semantic information have also been incorporated into manipulation through semantic grasp generation~\cite{liSemGraspSemanticGrasp2024}, task-aware grasp ranking~\cite{appiusTaskAwareRoboticGrasping2025}, SemSegGrasp~\cite{heSemSegGraspPlugandplayTaskoriented2025}, and VLM-based constraint generation such as ReKep~\cite{huangReKepSpatioTemporalReasoning2024}.
RoboSeg is complementary to these methods.
Its focus is the engineering bridge between online part-level semantic mapping and task-conditioned grasp selection: unlike methods that primarily reformulate task-oriented grasping as point-cloud semantic segmentation or direct grasp ranking, RoboSeg studies how an eye-in-hand robot constructs and stabilizes a persistent 3D part map before downstream grasp proposal.

\section{System Overview}

\begin{figure*}[t]
    \centering
    \includegraphics[width=\textwidth]{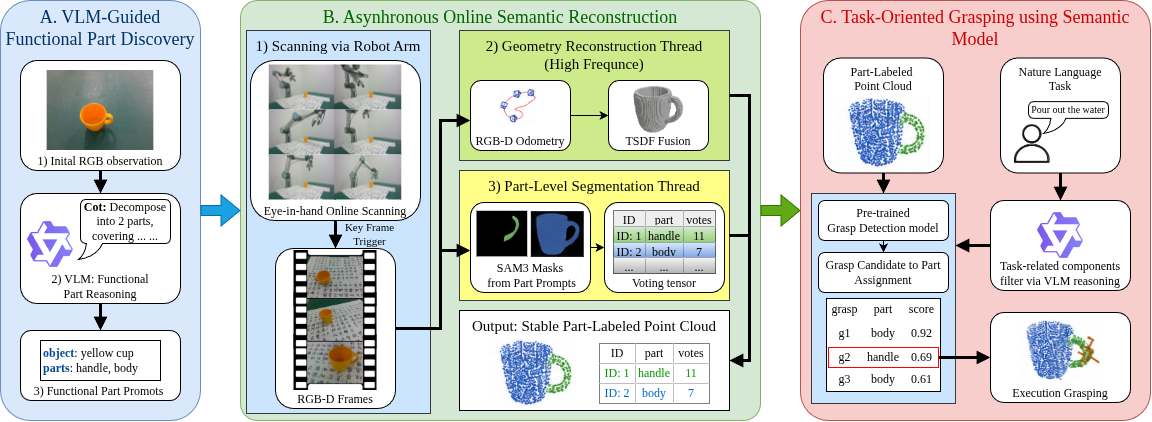}
    \caption{\textbf{RoboSeg system architecture.} \textit{A: VLM-guided functional part discovery} converts the initial RGB observation into an object name and compact functional part prompts. \textit{B: Asynchronous online semantic reconstruction} decouples high-frequency RGB-D odometry and TSDF fusion from keyframe-triggered SAM3 part-mask prediction; projected masks are accumulated in a voxel-voting tensor to produce a stable part-labeled point cloud. \textit{C: Task-oriented grasping} resolves a natural-language task to task-relevant part labels, assigns pretrained grasp-detector candidates to semantic parts, and executes the selected candidate.}
    \label{fig:system_overview}
\end{figure*}

RoboSeg is organized into three modules (Fig.~\ref{fig:system_overview}).
Module A performs VLM-guided functional part discovery: the robot captures an initial RGB observation, the VLM decomposes the visible object into one to three self-contained manipulation-relevant parts, and only those part names become SAM3 prompts.
Module B performs asynchronous online semantic reconstruction: eye-in-hand RGB-D frames drive a high-frequency geometry thread for odometry and TSDF fusion, while keyframes trigger a lower-rate part-segmentation thread whose projected masks are integrated by voxel voting.
Module C performs task-oriented grasping: a natural-language task is resolved to the relevant part label, AnyGrasp proposes 6-DoF candidates on the reconstructed point cloud, and each candidate is assigned the dominant semantic instance under the gripper closing region before task-conditioned selection.

Algorithm~\ref{alg:roboseg} summarizes a single run of the pipeline.
The A--C markers correspond to the modules in Fig.~\ref{fig:system_overview}: VLM-derived part prompts define the SAM3 vocabulary, keyframe masks update persistent voxel votes, and task-relevant part labels filter AnyGrasp candidates.

\section{Method}

\subsection{VLM-Driven Part Discovery}

The VLM constrains an existing segmentation model to manipulation-relevant outputs; without this constraint, SAM-style masks often follow printed patterns, shadows, seams, or highlights and create cluttered grasp targets.
RoboSeg queries Qwen3-VL~\cite{yangQwen3TechnicalReport2025} for part reasoning before scanning.
The prompt asks the model to name the object and list one to three functional parts that should be segmented for manipulation.
The required output format is
\begin{equation*}
\{\texttt{object}: o,\; \texttt{parts}: [p_1,\ldots,p_n],\; \texttt{\_cot}: r\},
\end{equation*}
where $p_i$ must be self-contained and unambiguous.
For example, \textit{handle} is replaced by \textit{watering can handle} when multiple objects may be visible.
Only the \texttt{parts} field is forwarded to SAM3~\cite{carionSAM3Segment2025}; the rationale is not used downstream.
This design reduces the semantic search space from arbitrary visible regions to a compact set of functional parts and limits the number of masks that must be fused in 3D.

\begin{algorithm}[t]
\caption{RoboSeg end-to-end pipeline}
\label{alg:roboseg}
\begin{pseudo}
\item \kw{Input:} initial RGB frame $I_0$, RGB-D stream $\{(I_t,D_t)\}$, task $T$, keyframe interval $\tau$
\item \kw{Output:} task-conditioned grasp $g^\star$ on the requested part
\item $(o,\{p_1,\dots,p_n\},r)\!\leftarrow\!\mathrm{VLM}(I_0)$; forward $P\!=\!\{p_i\}$ to SAM3\hfill$\triangleright$ A
\item init.\ empty TSDF volume and per-voxel vote tensors $\mathcal{S}(\cdot)$\hfill$\triangleright$ B
\item \kw{for} each RGB-D frame $(I_t,D_t)$ \kw{do}\hfill$\triangleright$ geometry thread
\item \IND estimate pose $\hat{T}_t$ (FK-regularized); fuse $D_t$ into TSDF
\item \IND \kw{if} $t \bmod \tau = 0$ \kw{then}\hfill$\triangleright$ semantic thread
\item \IND\IND $\mathcal{M}\!\leftarrow\!\mathrm{SAM3}(I_t,P)$; back-project mask pixels to voxels
\item \IND\IND \kw{for} each mask (label $\ell$, voxel set $\mathcal{V}_m$) \kw{do}
\item \IND\IND\IND $i\!\leftarrow\!$ same-class ID found in $\mathcal{V}_m$, \kw{else} allocate new ID
\item \IND\IND\IND \kw{for} $v\!\in\!\mathcal{V}_m$: add one vote for $i$; keep top-$K\!=\!3$ of $\mathcal{S}(v)$
\item \IND\IND \kw{end for}
\item \IND \kw{end if}
\item \kw{end for}
\item export cloud; per-voxel label $\hat{i}(v)\!=\!\arg\max_k s_k(v)$; build ID-to-part table\hfill$\triangleright$ C
\item $G\!\leftarrow\!\mathrm{AnyGrasp}(\text{reconstructed cloud})$
\item \kw{for} each candidate $g\!\in\!G$ \kw{do}
\item \IND assign $g$ to the dominant instance in its closing region
\item \kw{end for}
\item $L\!\leftarrow\!$ task-relevant part labels resolved from $T$ via the table
\item $g^\star\!\leftarrow\!$ highest AnyGrasp-score candidate whose label $\in L$
\item \kw{return} $g^\star$
\end{pseudo}
\end{algorithm}

\subsection{Asynchronous Online Semantic Reconstruction}

Module B in Fig.~\ref{fig:system_overview} converts the eye-in-hand RGB-D scan into a stable part-labeled point cloud.
The main design choice is to separate metric reconstruction from semantic inference: the robot should continue tracking and fusing depth frames even when SAM3 inference is slower than the camera stream.
RoboSeg therefore uses two asynchronous streams that share the same voxel coordinate frame.

\subsubsection{Asynchronous Semantic Tracking}

The high-frequency geometry thread estimates camera poses, optionally regularized by robot forward kinematics, and integrates every depth image into a TSDF volume.
In parallel, the semantic thread selects keyframes at a configurable interval, runs SAM3 with the VLM-generated part prompts, and back-projects valid mask pixels into the current TSDF voxel coordinates.
This decoupling keeps odometry and geometry fusion responsive while semantic labels arrive only when keyframe masks become available.

\subsubsection{Temporal Consensus via Voxel Voting}

The semantic stream must also convert noisy 2D part masks into persistent 3D labels.
Per-frame masks are not temporally stable: viewpoint changes can move boundaries, split a part, or match a neighboring region.
Naively overwriting voxel labels therefore creates flickering 3D semantics.
RoboSeg instead attaches a compact voting tensor to each occupied voxel.
For voxel $v$, the tensor stores the top-$K$ candidate instance IDs and their vote counts:
\begin{equation}
\mathcal{S}(v)=\{(i_1(v),s_1(v)),\ldots,(i_K(v),s_K(v))\},\quad K=3.
\end{equation}
When a new mask with class label $\ell$ projects to a voxel set $\mathcal{V}_m$, the system searches for an existing same-class instance ID in those voxels.
If no match exists, a new instance ID is allocated.
The selected ID receives one vote in every voxel covered by the mask.
At query time, the semantic label is determined by the maximum-vote instance,
\begin{equation}
\hat{i}(v)=\arg\max_k s_k(v).
\end{equation}
Transient 2D errors receive too few votes to dominate, while repeatedly observed part regions converge to stable labels.

\subsection{Task-Conditioned Grasp Candidate Selection}

After scanning, RoboSeg exports the reconstructed point cloud, per-point semantic instance IDs, and an instance table mapping IDs to part names.
The task-oriented grasping stage uses this part-labeled point cloud as a semantic model rather than as a replacement for a grasp detector.
AnyGrasp is applied to the reconstructed geometry to generate dense 6-DoF grasp candidates.
For each candidate, the system queries points inside the gripper closing region and assigns the candidate to the dominant semantic instance.
Given a natural-language task such as \textit{grasp the handle to lift the mug}, task-level reasoning maps the instruction to one or more part labels in the instance table; the selector then keeps candidates whose semantic label matches the requested part and ranks them by AnyGrasp score.
This stage does not attempt to improve the raw AnyGrasp confidence.
Instead, it converts a single score-ranked grasp list into part-indexed choices that can be constrained by a language-level task.
The design therefore treats the semantic map as an actionable interface for downstream manipulation, rather than only as a visualization.

\section{Experiments}

\subsection{Experimental Setup}

Experiments use a UR5 manipulator with an Intel RealSense D435i mounted in an eye-in-hand configuration.
The implementation combines RGB-D tracking and TSDF fusion, SAM3~\cite{carionSAM3Segment2025} segmentation, Qwen3-VL~\cite{yangQwen3TechnicalReport2025} part discovery, AnyGrasp~\cite{fangAnyGraspRobustEfficient2023} detection, task-part filtering, and MoveIt-compatible grasp execution.
Inference and mapping run on a workstation with an NVIDIA RTX 3090 GPU.
The evaluated objects are household and workshop items with manipulation-relevant parts, including mugs, bowls, kettles, tools, bottles, pots, and a watering can.

The experiments evaluate three questions: whether RoboSeg discovers manipulation-relevant parts during online scanning, whether the resulting maps align with annotated object geometry, and whether task-conditioned part labels improve grasp selection and support physical manipulation on a real robot.
Throughout this paper, \emph{online} denotes that the part-level map is constructed incrementally while the robot scans an unknown object, rather than from a pre-captured complete mesh or point cloud.

\subsection{Comparison with Offline Part Segmentation Methods}

The first experiment compares RoboSeg with representative offline 3D part segmentation methods under deliberately favorable input conditions for the offline baselines.
PartSLIP~\cite{liuPartSLIPLowShotPart2023,zhouPartSLIPEnhancingLowShot2023} and PartField~\cite{liuPARTFIELDLearning3D2025} are evaluated on the available object mesh or processed complete geometry, rather than on RoboSeg's reconstructed scan.
RoboSeg, in contrast, does not assume a pre-existing mesh: it builds its part-level representation from robot-acquired RGB-D observations during online scanning.
This comparison therefore asks whether an online manipulation system can recover functional parts at a quality comparable to, and in some cases more task-aligned than, offline methods that receive stronger geometric input.

\begin{figure}[t]
    \centering
    \includegraphics[width=0.48\textwidth]{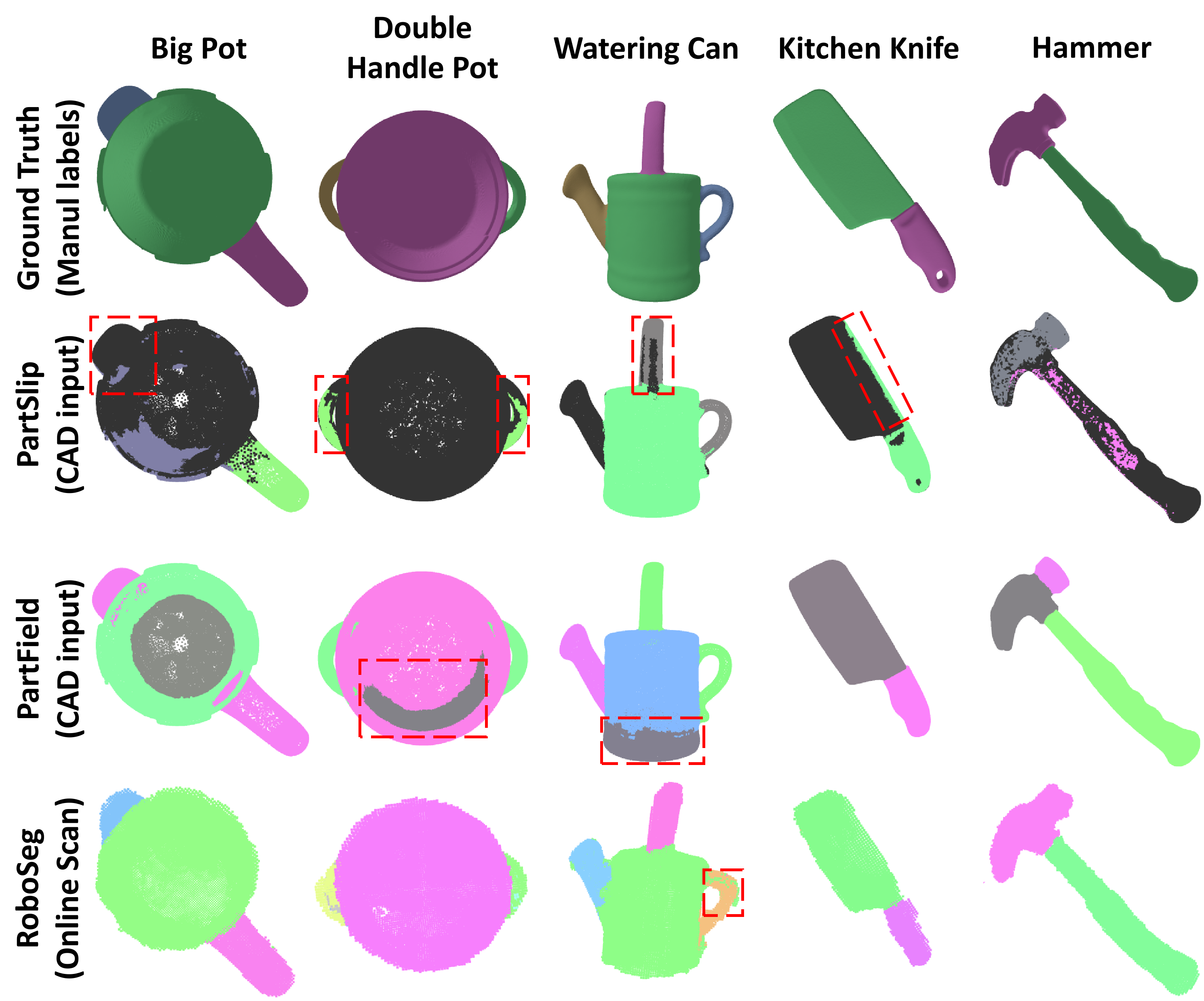}
    \caption{\textbf{Qualitative functional-part comparison.} Columns show Big Pot, Double Handle Pot, Watering Can, Kitchen Knife, and Hammer. PartSLIP and PartField operate on clean mesh or processed complete geometry, while RoboSeg constructs part labels from robot-acquired online scans without mesh input. Red dashed boxes highlight manipulation-critical regions such as handles, spouts, blades, and tool heads.}
    \label{fig:offline_comparison}
\end{figure}

Fig.~\ref{fig:offline_comparison} qualitatively compares functional part structure rather than reporting a standardized segmentation benchmark.
The offline methods receive complete geometry and can produce plausible object-level decompositions, but the resulting regions do not always isolate the parts required by manipulation tasks.
In the shown examples, PartSLIP often absorbs handles, spouts, blades, or tool heads into larger unlabeled or mixed regions, while PartField tends to separate broad geometric components without consistently aligning them with the functional part to be grasped or avoided.
RoboSeg infers and fuses part labels from partial online observations, yet separates the same task-critical regions more directly in these examples.
This comparison supports the claim that RoboSeg can obtain a manipulation-aligned part representation without relying on an offline mesh.

To quantify the visual comparison where aligned labels are available, we annotate functional parts for the five RoboSeg reconstructions in Fig.~\ref{fig:offline_comparison} and evaluate point-level IoU.
Using Blender material annotations as ground truth (GT) and ignoring prediction points farther than 5~mm from the GT surface, RoboSeg obtains 83.4\% mIoU across BigPot (85.8\%), DoubleHandlePot (61.9\%), Hammer (99.8\%), KitchenKnife (99.3\%), and WateringCan (70.1\%).
The lower scores stem from thin or protruding parts such as one pot handle and the watering-can spout.
Four aligned objects (BigPot, DoubleHandlePot, Hammer, KitchenKnife) also support a comparison with PartField's vertex-aligned segmentation under the identical merged material labels: under this common metric RoboSeg reaches 86.7\% mIoU on these objects, while PartField reaches 83.2\% (BigPot 76.5\%, DoubleHandlePot 71.8\%, Hammer 88.0\%, KitchenKnife 96.7\%).
This comparison uses shared labels but different input conditions: RoboSeg is evaluated on online-scan points projected to the GT surface with a 5~mm cutoff, whereas PartField is evaluated on the complete vertex-aligned mesh.
Under this restricted but shared label frame, RoboSeg remains comparable to PartField despite not using the complete mesh as input.
PartSLIP remains qualitative because its three matched outputs (BigPot, DoubleHandlePot, KitchenKnife) are not aligned under this metric frame, and Hammer/WateringCan are not available for the same comparison.

\subsection{Geometric Accuracy of RGB-D Odometry}

The second experiment isolates the geometric backbone.
RGB-D odometry and TSDF fusion are evaluated against aligned CAD or dataset geometry for representative objects with reference models.
For each object, we compare two pose sources for TSDF fusion: robot forward-kinematic poses recorded during data collection and poses estimated by RGB-D odometry.
The comparison is paired by object identity, with RGB-D-odometry rows taken from the aligned online scan for each object.
Metrics are computed on the semantic instance cloud rather than the full TSDF cloud, because the instance cloud removes the support plane during SAM3-guided fusion.
The metrics are Chamfer distance, reconstruction-to-ground-truth accuracy, ground-truth-to-reconstruction completeness, and F1 scores at 5~mm and 10~mm.

\begin{table}[t]
\centering
\caption{Aggregate geometry metrics by pose source.}
\label{tab:geometry_metrics_compact}
\scriptsize
\setlength{\tabcolsep}{2.5pt}
\begin{tabular}{@{}lrrr@{}}
\toprule
Metric & Kinematic & RGB-D odom & $\Delta$ \\
\midrule
Cham. $\downarrow$ (mm) & 11.1$\pm$7.8 & 10.7$\pm$10.9 & -0.4 \\
Acc. $\downarrow$ (mm) & 4.9$\pm$3.6 & 4.6$\pm$3.3 & -0.3 \\
Comp. $\downarrow$ (mm) & 6.3$\pm$8.0 & 6.2$\pm$7.7 & -0.1 \\
F1@5 $\uparrow$ (\%) & 74.3$\pm$20.7 & 75.0$\pm$25.7 & +0.7 \\
F1@10 $\uparrow$ (\%) & 88.9$\pm$16.2 & 87.2$\pm$20.7 & -1.7 \\
\bottomrule
\end{tabular}
\vspace{1mm}\par\footnotesize Values are mean$\pm$std over five paired objects. Delta is RGB-D odometry minus kinematic pose integration.
\end{table}

Table~\ref{tab:geometry_metrics_compact} reports aggregate behavior over objects with aligned CAD or dataset geometry.
Kinematic-pose rows use robot forward kinematics, while RGB-D-odometry rows use visually estimated camera poses from the same aligned scans.
The pose-source effect is mixed: RGB-D odometry is slightly better on Chamfer distance, accuracy, completeness, and F1@5, but lower on F1@10.
Hammer is the largest source of variance, and KitchenKnife remains difficult because thin geometry makes completeness and F1 sensitive to small alignment errors.

\subsection{Pilot Physical Task-Part Grasp Validation}

This experiment tests whether RoboSeg's part-labeled map can serve as an execution interface for task-conditioned grasping.
The experiment isolates semantic routing on top of AnyGrasp proposals, rather than introducing a new grasp generator or an executed comparison against a part-agnostic selector.
The key distinction is between physical grasp feasibility and task-conditioned semantic correctness.
Physical success requires the robot to grasp the object, lift it, and maintain a stable hold during the commanded lift.
Task-part success requires the system to resolve the instruction to the intended functional part and execute a grasp that contacts that part, independent of whether the subsequent lift remains mechanically stable.
The combined success metric therefore requires both conditions: the robot must grasp the requested part and physically lift the object without failure.

For each trial, RoboSeg scanned the scene once from an informative initial view, resolved the task instruction to a target semantic part, generated AnyGrasp 6-DoF candidates on the reconstructed cloud, and executed the highest-ranked candidate whose contact point lay on the specified part.
The evaluation contains 24 physical trials over the eight object-task conditions listed in Table~\ref{tab:level3_physical_grasp_summary}; each condition was repeated three times with manual reset.
We further analyze recorded AnyGrasp proposal sets for the same tasks.
Target-part candidates are present in all 8/8 conditions, and RoboSeg's semantic filtering selects a target-part candidate in all 8/8 conditions; by contrast, the unfiltered AnyGrasp top-1 candidate matches the target part in only 4/8 conditions.
This proposal analysis re-ranks recorded candidates only and is therefore reported separately from the physical trials.
To assess candidate-indexing coverage more broadly, we also measure how often reconstructed part labels index generated proposals.
Across 21 reconstructed scenes with AnyGrasp outputs, the candidate-to-part assignment step labels 792 of 883 proposals (89.7\%) with a known functional part, showing that the reconstructed map covers most generated grasp candidates.

\begin{table}[t]
\centering
\caption{Pilot physical task-part grasp validation.}
\label{tab:level3_physical_grasp_summary}
\scriptsize
\setlength{\tabcolsep}{1.5pt}
\renewcommand{\arraystretch}{1.08}
\begin{tabularx}{\linewidth}{@{}lYlrrrr@{}}
\toprule
Object & Task instruction & Target & Trials & Phys. & Part & Both \\
\midrule
BigPot & Lift the pot by the main body & main body & 3 & 3/3 & 3/3 & 3/3 \\
BigPot & Hand the pot to a person & side handle & 3 & 3/3 & 3/3 & 3/3 \\
Hammer & Hold the red handle to drive a nail & red handle & 3 & 3/3 & 3/3 & 3/3 \\
Hammer & Hand over the hammer to a person & hammer head & 3 & 3/3 & 3/3 & 3/3 \\
KitchenKnife & Hand the kitchen knife to a person for cutting & blade & 3 & 1/3 & 3/3 & 1/3 \\
WateringCan & Prepare to water the plants & handle & 3 & 3/3 & 3/3 & 3/3 \\
WateringCan & Lift the watering can by the spout & spout & 3 & 2/3 & 3/3 & 2/3 \\
WateringCan & Hand the watering can to a person for use & handle & 3 & 3/3 & 3/3 & 3/3 \\
Overall & 8 object-task conditions & -- & 24 & 21/24 & 24/24 & 21/24 \\
\bottomrule
\end{tabularx}
\vspace{1mm}\par\footnotesize Phys. is physical grasp success; Part is whether the executed grasp contacted the intended task part; Both requires both.
\end{table}

Table~\ref{tab:level3_physical_grasp_summary} reports task-level results.
The system achieved 21/24 physical successes, 24/24 task-part successes, and 21/24 combined task successes.
The 24/24 task-part success count indicates that the semantic map and filtering stage consistently routed execution to the requested functional part in this pilot.
The lower physical and combined success counts attribute the remaining failures to grasp execution rather than semantic selection: the three failed trials still contacted the correct part, but did not produce a stable lift.
Because the pilot is small, these counts should be read as feasibility evidence rather than a precise estimate of physical success rate.
These results support the claim that RoboSeg provides a usable semantic indexing interface for task-conditioned grasp execution.

\section{Discussion and Conclusion}

RoboSeg integrates VLM functional part discovery, asynchronous online RGB-D semantic mapping, and AnyGrasp candidate indexing to convert part-level maps into task-conditioned grasp candidates on real objects, without requiring CAD models, object templates, or pre-scanned meshes.
The evaluation supports this bounded systems claim through physical pilot trials, pose-source comparisons, qualitative comparison with offline part segmentation methods, semantic-IoU evaluation on manually labeled objects, and grasp-candidate assignment analysis.
These results show that the reconstructed part map can serve as a semantic indexing interface for manipulation while AnyGrasp remains the proposal generator.
The current study is limited by 24 physical trials with manual reset and standardized initial views, a small set of objects with compatible manual annotations, incomplete metric-frame alignment for all offline baselines, and the absence of component-level runtime profiling.
Future work should expand physical trials, evaluate robustness under broader object poses and occlusions, extend semantic evaluation with reliably aligned baselines, profile runtime at component level, and compare against additional robot-executed selection policies.

\bibliographystyle{IEEEtran}
\bibliography{ref}

\end{document}